# Coverage based testing for V&V and Safety Assurance of Self-driving Autonomous Vehicles: A Systematic Literature Review

Zaid Tahir
Department of Computer Science
University of York
York YO105GH, UK
zaid.tahir@york.ac.uk

Rob Alexander
Department of Computer Science
University of York
York YO105GH, UK
rob.alexander@york.ac.uk

*Abstract* – Self-driving Autonomous Vehicles (SAVs) are gaining more interest each passing day by the industry as well as the general public. Tech and automobile companies are investing huge amounts of capital in research and development of SAVs to make sure they have a head start in the SAV market in the future. One of the major hurdles in the way of SAVs making it to the public roads is the lack of confidence of public in the safety aspect of SAVs. In order to assure safety and provide confidence to the public in the safety of SAVs, researchers around the world have used coverage-based testing for Verification and Validation (V&V) and safety assurance of SAVs. The objective of this paper is to investigate the coverage criteria proposed and coverage maximizing techniques used by researchers in the last decade up till now, to assure safety of SAVs. We conduct a Systematic Literature Review (SLR) for this investigation in our paper. We present a classification of existing research based on the coverage criteria used. Several research gaps and research directions are also provided in this SLR to enable further research in this domain. This paper provides a body of knowledge in the domain of safety assurance of SAVs. We believe the results of this SLR will be helpful in the progression of V&V and safety assurance of SAVs.

*Index Terms* – Safety assurance, V&V, self-driving cars, autonomous vehicles, coverage criteria.

## I. INTRODUCTION

The topic of Self-driving Autonomous Vehicles (SAVs) has been the subject of interest of a plethora of research in the past decade and currently is one of the main focus of the industry, academia and even governments as a whole. Self-driving cars are being tested on our roads today like Waymo [1], Uber [2], Tesla [3], etc. The predicted benefits of SAVs for the people are many e.g. an ease in commuting without the stress of driving, ample time available to do other tasks while commuting, eradicating the dependency of older people on young ones to travel in their own cars, pick and drop of children from schools without the need of a dedicated driver, to solve the last mile problem, etc. Also a few of the main goals for the development of SAVs are to save lives by reducing the number of traffic accidents, reducing traffic jams to save time for everyone, reducing carbon emissions for a greener tomorrow and many more. In light of these anticipated benefits of SAVs, one can imagine tremendous resources, lives and money being saved. Hence, people are ready to make huge investments in these SAVs [4]. A recent study from Allied Market Research [5] suggests that the market for self-driving cars will increase from 54.23 billion USD in 2019 up to 556.67 billion USD by 2026 with a 39.47% compound annual growth rate during that time span.

From a safety engineering perspective, the safety-critical nature of SAVs makes us question that if it is even possible to achieve a future with less traffic casualties by deploying SAVs. In 2016 a report published by National Highway Traffic Safety Administration (NHTSA) stated that 94% of the total 37,461 fatalities in traffic accidents were due to human error [6]. If the hazards associated with SAVs can be dealt with and in turn the traffic fatalities can be reduced, then SAVs do carry the potential to have an impact on human lives.

SAV is an integration of a variety of complex systems to form a System of Systems (SoS), so it can come across a countless number of hazardous situations either from the outside environment or from within its SoS. Although original equipment manufacturers (OEMs) have tried their best to take care of these hazards during every phase of the development-cycle, yet there have been a few fatal accidents involving SAVs [7]. The SAVs involved in these fatal accidents were SAE level-3[1] or lower. While there have been only a few fatal crashes involving SAVs, many non-fatal accidents have taken place [8]. The diverse set of hazards faced by a SAVs make verification and validation (V&V) for the safety assurance of SAVs an extremely challenging task.

Researchers have tried to tackle this problem of safety assurance of SAVs with different strategies. There is no standard testing procedure for V&V of SAVs and neither do the traditional software testing methods and practices for testing traditional cars translate effectively to SAVs [9]. One of the reasons for the difficulty in testing SAVs is that the number of inputs for testing is so large as to be effectively infinite, when considering the external and internal hazards.

In software testing one method used to cover the huge number of possible inputs effectively is coverage criteria [10]. Coverage criteria provides a practical and structured approach to search the input space, to decide which inputs to use for testing and when to stop testing [10]. With the upside of coverage criteria in mind, recently many researchers have used coverage-criteria-based testing for the V&V and safety assurance of SAVs (which we will now on refer to as coverage-based V&V and safety assurance of SAVs).

---

[1]Society of Automotive Engineers (SAE) has developed six levels of driving automation according to SAE J-3016 international report at https://www.sae.org

In order to develop a body of knowledge of this intricate yet nascent field, we decided to conduct a systematic literature review (SLR) on this topic of coverage-based V&V and safety assurance of SAVs, which as per authors knowledge, is novel and would prove to be very useful for the progression of V&V and safety assurance of SAVs. As a part of this study, a classification for coverage criteria used for V&V and safety assurance of SAVs was developed which is discussed in detail in the following sections.

The objective of this paper is threefold: (1) To conduct a SLR of the chosen topic with the aim of answering the most pertinent Research Questions (RQs) and highlighting key challenges and research gaps; (2) Discuss possible future research directions; (3) To provide a background allowing new research an early start with a strong foundation.

The remainder of this papers is divided into following sections. Section II presents the SLR research methodology along with the proposed RQs. Section III follows up with detailed analysis and results of the SLR as per the RQs. Future research directions and research gaps are identified in Section IV. Section V concludes the paper followed by acknowledgements and references.

## II. RESEARCH METHODOLOGY

### A. Research Questions

According to Goal Question Metric (GQM) we define our free-form question as "Discover and evaluate the scientific contributions on coverage-based V&V and safety assurance of SAVs." This question is broken down into further RQs exploring key issues of coverage-based V&V and safety assurance of SAVs. Following the guidelines presented in [11], [12], [18], the RQs considered for this SLR are presented below:

*1) RQ1*: How is the SAV coverage-based V&V evolving in the last decade in terms of geographic distributions and its growth over the years? Having answer to this RQ1 will help readers highlight active labs researching on this topic, hence make it easier for them to find new contributions in this area and collaborate with those labs for further research.

*2) RQ2*: Is the coverage criteria explicitly mentioned or do we have to implicitly derive the coverage criteria out of the studied paper? V&V of SAVs is a primordial state-of-the-art field and consequently people all around the world use different terms for a single technique. That's why in many papers, the authors do not explicitly mention what coverage criteria they are using. We are going to infer the implicit coverage criteria the authors are using in the respective studied papers in which coverage criteria isn't explicitly mentioned. The answer to this question RQ2 would be very useful to the readers as it will give a distribution of the papers that are explicitly using this terminology of coverage criteria as a medium to convey confidence in the V&V of SAVs and how many papers are implicitly using this term, thereupon providing a better understanding of coverage-based V&V and safety assurance of SAVs.

*3) RQ3*: What coverage criteria are being used in the studied papers? We consider this question to be very important, as the market for SAVs is expected to grow, this classification of studied publications based on coverage-based V&V and safety assurance of SAVs will give a bird's eye view to the readers as to what metrics have been used up till now for conveying confidence in the V&V of SAVs.

*4) RQ4*: What is the technique being used to ensure maximum coverage of the coverage criteria under consideration of each studied paper i.e. Random number generation, Search-Based Software Testing (SBST), Machine Learning (Deep Learning, Reinforcement Learning), etc.? Answer to this question will give readers cues to what techniques are currently being used and what techniques haven't been tried yet for achieving maximum possible coverage, in turn opening up new research avenues.

*5) RQ5*: What is the testing set up used to execute the proposed experiments in the papers? Is simulation/virtual testing used or is there hardware involved as well or is there some other method such as on paper proofs? The answer to this RQ5 will give a guideline to the readers for choosing their testing set up.

### B. Review Policy

The review policy as depicted in Fig. 1, follows the formulation of RQs. The review policy involves first selecting our data resources for searching research related to our SLR. Then the inclusion/exclusion criteria are selected by the reviewers regarding which papers to include and exclude from our searches of the research data resources till the stop criterion is reached. The Quality Criteria (QC) is then defined by the reviewers to rate the quality of the reviewed research. The following subsections elaborate on the review policy.

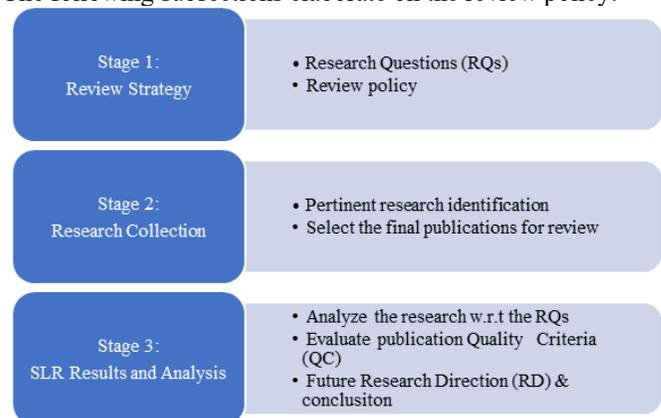

Fig. 1 SLR Research Methodology.

### C. Research Data Resources

The research data resources were selected based on their relevance to the topic of this SLR. The selected research data resources are as follows: (I) IEEE Xplore; (II) ACM Digital Library; (III) SpringerLink; (IV) Scopus; (V) Google Scholar. Google Scholar was specifically chosen as it provides links to both peer reviewed and non-peer reviewed articles in our searches which can prove to be very useful, as this field of coverage-based V&V and safety assurance of SAVs is new and a lot of valuable research is being published in various forms including and excluding peer reviewed articles e.g. Masters, PhD thesis and articles from open access repositories such as Arxiv, etc. After the selection of research data resources, the search string/keywords sets for searching the research data base were developed based on the authors' knowledge of the topic of this SLR and iteratively searching the research data resources till the search string sets were refined. It is worth noting that there were query words limits in the various research data resources as certainly these data resources can't query infinite words due to limited processing power available e.g. there was a limit of 32 query words on Google Scholar. In the end we ended up with a couple of search string sets which are as follows: (I) *{("safety assurance" OR "verification" OR "validation" OR "V&V" OR "verification and validation" OR "coverage" OR "testing") AND ("autonomous vehicle" OR "autonomous system" OR "self-driving car" OR "driverless car")}*; (II) *{("situation" OR "scenario" OR "requirements") AND ("coverage" OR "testing" OR "safety assurance" OR "verification" OR "validation" OR "V&V" OR "verification and validation") AND ("autonomous vehicle" OR "autonomous system" OR "self-driving car" OR "driverless car")}*. These search string sets were used to search within the titles, keywords, abstracts and full text of publications. This was done to make sure maximum articles are acquired, as research on coverage-based V&V and safety assurance of SAVs is scarcely available as compared to mature fields e.g. control systems.

The research data resources after being queried with the search string sets returned a substantially large set of articles, especially resources such as IEEE Xplore, Google Scholar and ACM Digital Library. Many of the articles in this large set returned by these resources weren't relevant to the topic of this SLR and to the RQs specified previously. In order to trim down the irrelevant articles and limit our search we used the following stop criterion as adopted in [13]: "*Stop the collecting of articles after a sequence of 10 titles, completely incoherent with the query, appeared in the list.*" The incoherence of the articles as mentioned in this stop criterion is determined by the authors.

### D. Inclusion Exclusion Criteria

After collection of articles from the specified research data resources using the search string sets and applying the established stop criterion, we obtain a pool of articles which may still contain many articles which don't answer our RQs sufficiently. On these grounds it appears that most SLRs [11], [14], [15] use inclusion exclusion criteria to trim down the pool of articles to only the pertinent ones. The inclusion & exclusion criteria have been defined below:

*1) Inclusion criteria*: Following are the inclusion criteria:

*In1) Related to topic*: Studies related to coverage-based V&V and safety assurance of SAVs

*In2) Autonomous systems*: Studies related to coverage-based V&V and safety assurance of autonomous systems are included, even if they are not specifically SAVs e.g. industrial fleet of autonomous robots, etc. The reason for this inclusion of autonomous systems is that there are very few papers that directly address this issue of coverage-based V&V and safety assurance of SAVs, furthermore the ideas and concepts used for coverage-based V&V and safety assurance of autonomous systems apply directly or indirectly to SAVs as well because SAVs are a subclass of autonomous systems.

*In3) Last date of collection*: Studies published up till August 2019. As the SLR research data collection was done up till this date.

*In4) Type of articles*: Peer reviewed as well as non-peer reviewed research e.g. dissertations, technical reports etc.

*2) Exclusion criteria*: Exclusion criteria are as follows:

*Ex1) Invalid type of article*: Examples include poster, demo, presentation, catalogue, etc. Such articles are excluded.

*Ex2) Publication Date*: The publications that date back more than a decade are excluded. As the technology rapidly evolves, the old technology tends to become obsolete especially in case of SAVs.

*Ex3) Written language*: If the research article is not written in English, it is excluded.

*Ex4) Duplicate papers*: Redundant versions of the same paper are excluded.

*Ex5) Irrelevant research*: Studies whose area of research is not related to coverage-based V&V and safety assurance of SAVs or of autonomous systems are excluded.

*Ex6) Algorithms for operations and functionality of SAVs*: Research articles focused at developing algorithms for SAVs for their proper functionality are excluded. These articles are excluded since we are only interested in V&V and safety assurance of SAVs based on some coverage criteria.

*Ex7) SAE level 2 and below*: For the review we are considering mostly SAE level-4, level-5 and in some cases level-3 SAVs as the term "self-driving" generally refers to these. Publications referring to SAE level 2 and below are excluded.

The inclusion exclusion criteria are applied in two stages, a strategy adopted from [16]. First stage is coarse-grained inclusion exclusion stage, in which articles are excluded if even one of the exclusion criteria is satisfied by the title or abstract of that article. After coarse-grained inclusion exclusion, fine-grained inclusion exclusion is applied on the papers that pass through the first stage screening. In this stage, whole text of the article under observation is reviewed and is included or excluded depending upon its conformance with the inclusion exclusion criteria defined above.

### E. Quality Criteria for Quality Assessment

The quality assessment helps to make the conclusions drawn in this research more reliable and ascertains the coherent synthesis and credibility of the results [4]. That is why most SLRs rely on QC for quality assessment, as recommended by [11]. QC typically gives the measure whether the authors in the articles under review provide a clear statement of goals and overall a sound rationale. Following the guidelines of [11], [18], the quality assessment of the literature under review was achieved by a scoring technique with a score of 5 being the highest and a score of 1 being the lowest score for each QC. A set of six QC were used to evaluate all the papers. Consequently, we have done quality assessment by averaging the quality score of the QC for all the article reviewed. The QC are presented in Table I.

TABLE I

| ID | Quality Criteria |
|---|---|
| Q1 | Is there a clear statement of the goals of the research [17]? |
| Q2 | Is the proposed technique clearly described [17]? |
| Q3 | Is there an adequate description of the context (industry, laboratory setting, products used and so on), in which the research was carried out [17]? |
| Q4 | Are the limitations of this study explicitly discussed [2]? |
| Q5 | Was the proposed technique evaluated rigorously? |
| Q6 | Are the proposed coverage-based testing techniques potentially applicable and practical for AV industry? (Does it scale?) |

TABLE II
RESULTS: COARSE-GRAINED STAGE

| Research articles database | Number of articles | Percentage |
|---|---|---|
| Total research articles | 125 | 100% |
| Total articles excluded | 36 | 29% |
| Total articles included | 89 | 71% |

## III. REVIEW AND RESULTS ANALYSIS

In this section the actual review is performed. First the results of inclusion and exclusion stages are discussed. Both the steps of exclusion criteria i.e. coarse-grained exclusions and fine-grained exclusions are evaluated in detail in this section. Then an in-depth analysis of the SLR results are provided with respect to the RQs.

### A. Research Articles Selection

This subsection describes the final selection of research articles after their collection from the research data resources. The two search string sets thoroughly specified in the previous section, were used to search for the relevant articles in the research data resources. A total of 125 articles were collected from all the research data resources using the search sting sets with the application of the chosen stop criterion. Note that the number is not that large due to the fact that this field is in its nascent stages. Then coarse-grained inclusion exclusion is applied and as mentioned only the abstract and title of the papers are considered for exclusion. As depicted in Table II, 71% (89 papers) of total number of articles were included, referenced in [19], and 29% (36 papers) were excluded in the coarse-grained inclusion exclusion stage. What can be said about the publications included after the coarse-grained inclusion exclusion stage is that, these papers are related to V&V and safety assurance of SAVs, but not all are necessarily coverage-based V&V and safety assurance of SAVs, as only the title and abstract of the articles were examined in the coarse-grained inclusion exclusion stage, which is not enough to determine whether some coverage criteria was used or not. The geographical distribution of the publications obtained after the coarse-grained inclusion exclusion stage has been depicted in Fig. 2. USA tops the geographical distribution with the maximum number of papers related to V&V and safety assurance of SAVs, followed by European countries like Germany, UK, Italy, etc. Even though fine-grained inclusion exclusion hasn't been applied yet, still a trend can be construed from this distribution that those countries which have a larger share in the automobile industry across the globe, have the higher number of papers related to V&V of SAVs, which makes sense because these countries would be looking to introduce SAVs in the future. This trend will be further analysed after the fine-grained inclusion exclusion stage while answering RQ1 so that a clearer picture can be drawn about which countries have published related to coverage-based V&V and safety assurance of SAVs.

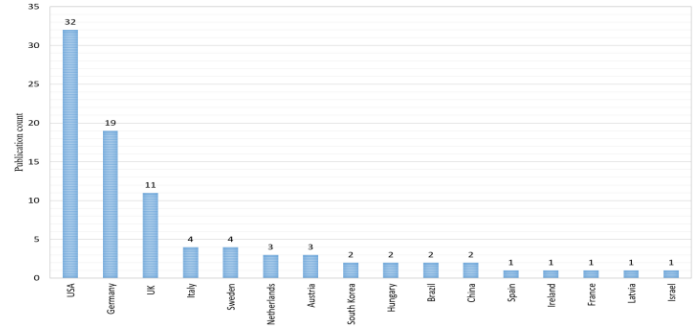

Fig. 2 Publications geographical distribution after coarse-grained stage.

Fig. 3 illustrates the number of papers after coarse-grained inclusion exclusion related to V&V and safety assurance of SAVs each year in the last decade. It can be seen that the number of papers published till 2015 is fairly low but a sudden spike in total papers published after 2015 can be seen, starting from 2016. The papers in 2019 are still higher than 2015 considering this SLR was performed for papers till August 2019, hence it is expected this number will rise till the end of 2019 taking into account the past trends. This trend in rise of papers related to V&V and safety assurance of SAVs in 2016 and the subsequent years, can be directly correlated to rise in public usage and demand of SAE level-3 SAVs and public road testing of SAVs for development of SAE level-4 and SAE level-5 SAVs of the top companies like Waymo [1], Uber [2], Tesla [3], etc. This sharp rise of research related to V&V and safety assurance of SAVs in 2016 and following years can also be attributed to the fact that the first fatal autonomous car crash in USA happened in 2016.

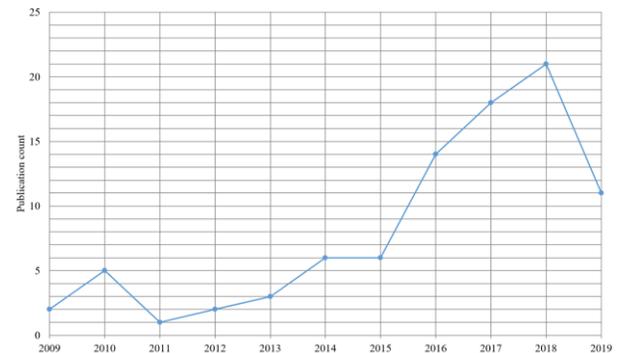

Fig. 3 Publications per year (till August 2019) after coarse-grained stage.

After coarse-grained inclusion exclusion stage comes fine-grained inclusion exclusion in which whole text of the article under observation is reviewed and the article is included or excluded depending upon the inclusion exclusion criteria. The article under observation is excluded if among the seven exclusion criteria, even one is satisfied. The results of the fine-grained inclusion exclusion are discussed in the following subsections to answer the RQs. It is important to note here that all RQs are answered according to the publications set acquired after fine-grained inclusion exclusion stage. The research works purely addressing the topic of this SLR i.e. coverage-based V&V and safety assurance of SAVs, can answers the proposed RQs and are only acquired after fine-grained inclusion exclusion stage. These articles obtained after fine-grained inclusion exclusion stage are listed along with their references in Table IV.

*B. Chronological & Geographical Distribution (RQ1)*

The answer related to chronological and geographical distributions of the literature pertaining to coverage-based V&V and safety assurance of SAVs has been answered in this subsection. Fine-grained inclusion exclusion was applied on the 89 articles that were acquired after coarse-grained inclusion exclusion stage. As depicted in Table III, 29% (26 papers) of the total number of articles (89 papers) were included and 71% (63 papers) were excluded after the fine-grained inclusion exclusion stage. It is clearly visible that a big percentage (71%) of papers were excluded in this fine-grained inclusion exclusion stage as compared to percentage of papers excluded in the previous coarse-grained inclusion exclusion stage (29%). Most of the articles excluded in this stage were due to exclusion criterion Ex5, as there were many papers related to V&V and safety assurance of SAVs but many of them didn't use coverage-based testing for it.

TABLE III
RESULTS: FINE-GRAINED STAGE

| Research articles database | Number of articles | Percentage |
|---|---|---|
| Total coarse-grained research articles | 89 | 100% |
| Total articles excluded | 63 | 71% |
| Total articles included | 26 | 29% |

Fig. 4 presents the geographical distribution of the publications acquired after the fine-grained inclusion exclusion stage.

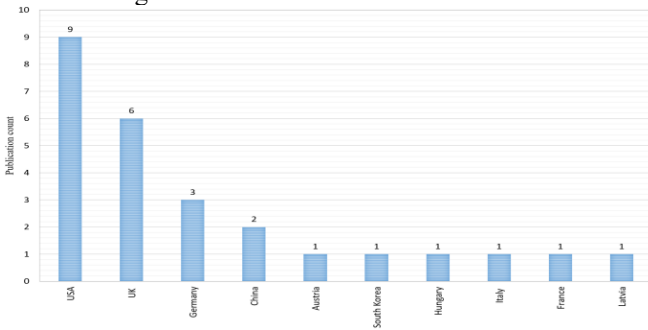

Fig. 4 Publications geographical distribution after fine-grained stage.

USA tops the countries with the greatest number of papers which is understandable with the allocation of funds and the top tech companies of SAVs such as Waymo [1], Uber [2], Tesla [3], etc., located in USA.

To understand the evolution of coverage-based V&V and safety assurance of SAVs we look at the related publications in the last decade obtained after fine-grained inclusion exclusion stage, illustrated in Fig. 5.

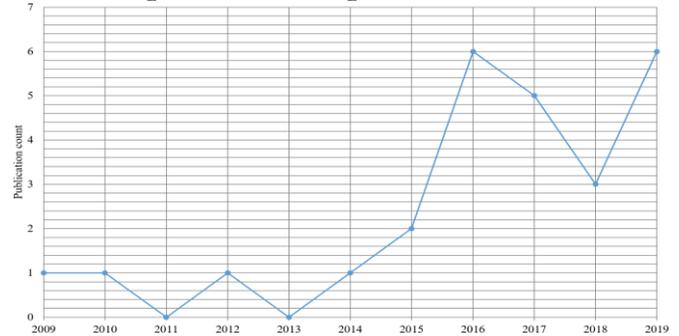

Fig. 5 Publications per year (till August 2019) after fine-grained stage.

As seen in Fig. 5, the number of papers started rising after 2015. The year 2019 till August, has the highest number of papers related to coverage-based V&V and safety assurance of SAVs and we assume that the number of papers will rise even more till the end of 2019, which directly points to the increasing importance of coverage-based V&V and safety assurance of SAVs as market of SAVs grows day by day.

*C. Explicit or Implicit Coverage Criteria (RQ2)*

After analyzing the publications obtained after fine-grained inclusion exclusion, it was our finding, as depicted in Fig. 6, that 46% of the total papers (12 out of 26) did not explicitly mention that they were using some sort of coverage criteria, and rest of the papers (14 out of 26) did mention the coverage criteria explicitly.

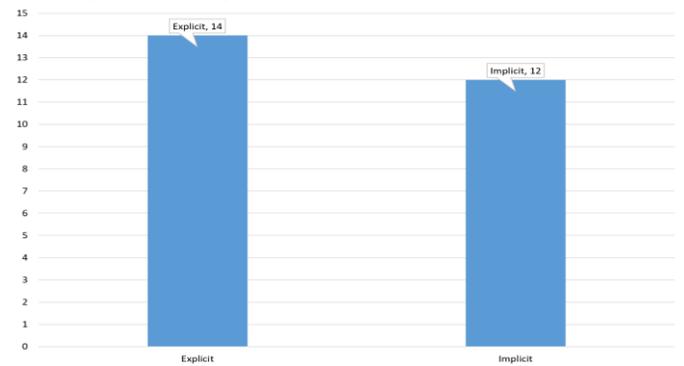

Fig. 6 Implicit vs explicit use of coverage criteria.

*D. Coverage Criteria Used (RQ3)*

There are different coverage-based testing approaches as mentioned in [27]. Upon detailed analysis of the set of articles acquired after fine-grained inclusion exclusion, we discovered that there were three kinds of coverage criteria being used. Each coverage criterion is explained below:

*1) Scenario coverage*: The term scenario has been defined by [20], in which the authors defined the term scenario as temporal development between scenes. Scenario coverage considers linear evolution from one scene to another e.g. autonomous vehicles changing lanes, following another vehicle, etc., and to cover the proportion of those pre-defined set of scenarios is scenario coverage.

*2) Situation coverage*: This term situation was widely adopted and used by Rob A. at el. [27] and it takes into account the different situations/elements internal or external an autonomous vehicles/systems can face and test them with those unexpected or expected situations, to cover the proportion of the situation space the SAV can face, is situation coverage.

*3) Requirements coverage:* In requirements coverage testing system under test (SUT) is assessed according to a set of identified requirements acceptability (according to a defined criterion of acceptability) [27].

The answer to the RQ3 generated by this SLR is seen in Fig. 6. Requirements coverage was the most used coverage criterion in all the publications related to coverage-based V&V and safety assurance of SAVs. 46% of the total publications (12 out of 26) used requirements coverage. 31% of the total publications (8 out of 26) used scenario coverage as the coverage criterion. The least used coverage criterion was situation-coverage, standing at 23% of the total publications (6 out of 26).

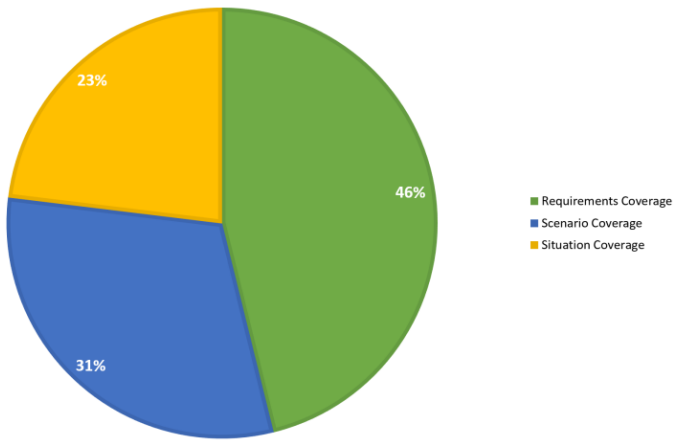

Fig. 6 Coverage criteria used in papers.

### E. Techniques for Maximizing Coverage (RQ4)

The results of this SLR showed that several techniques have been used by research works to maximize the coverage of the coverage criteria adopted for the V&V and safety assurance of SAVs. Fig 7 shows the distribution of techniques used to maximize the coverage during testing in the publications analyzed after fine-grained inclusion exclusion stage. From Fig. 7 we see that the techniques used to maximize coverage of the coverage criteria adopted by the research works can be classified into the following four categories: (1) Pseudorandom generation; (2) SBST; (3) Reinforcement learning; (4) Hight Throughput Testing (HTT). Fig. 7 shows that maximum research works use pseudorandom generation to maximize coverage (12 publications), followed by SBST (10 publications). Reinforcement learning comes at second last, being used in only three publication for maximum coverage and HTT was used in only one publication to maximize coverage.

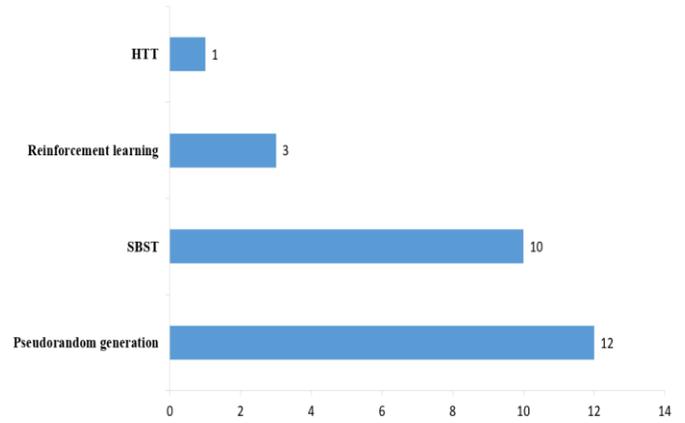

Fig. 7 Techniques used in publications for maximizing coverage.

Table IV lists each paper, obtained after fine-grained inclusion stage, as per the technique used for maximizing the coverage of the coverage criteria adopted for the V&V and safety assurance of SAVs.

TABLE IV
REVIEWED SLR PAPERS AS PER TECHNIQUE USED

| Techniques | Papers using this technique |
|---|---|
| Pseudorandom Generation | [21][22][6] [23][24][25][26][27][28][29][30][31] |
| SBST | [32][33][34][35][36][37][38][39][40][41] |
| Reinforcement Learning | [42][43][44] |
| HTT | [45] |

### G. Testing Setup (RQ5)

The purpose of this RQ is to evaluate the testing setups used in the research works under observation. To identify those that use simulations for their experimentations, which use hardware, which use both hardware and simulation for the experimentation of their proposed coverage-based V&V and safety assurance of SAVs.

The answer to this question has been given by this SLR and is plotted in Fig. 8.

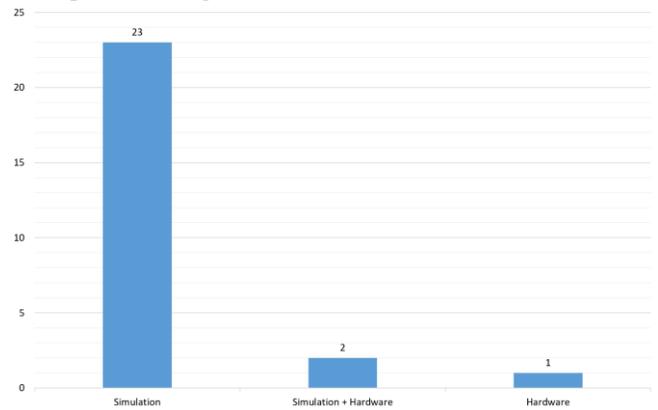

Fig. 8 Test setup used in publications.

Fig. 8 shows that a high number of research works rely only on simulating their experiments (23 papers) while one publication relies on hardware only and two publications use both simulation and hardware for the experimentation of their proposed coverage-based V&V and safety assurance of SAVs.

*H. Quality Assessment*

The QC presented in Table I were used to evaluate each publication obtained after fine-grained inclusion exclusion. Due to limited space we present the average quality assessment of all the publications for each QC in this SLR in Fig. 9. Lowest score being zero and highest score as five for each QC. It can be observed from Fig. 9 that QC1, QC2, QC3 have relatively higher scores as compared to QC4, QC5, QC6.

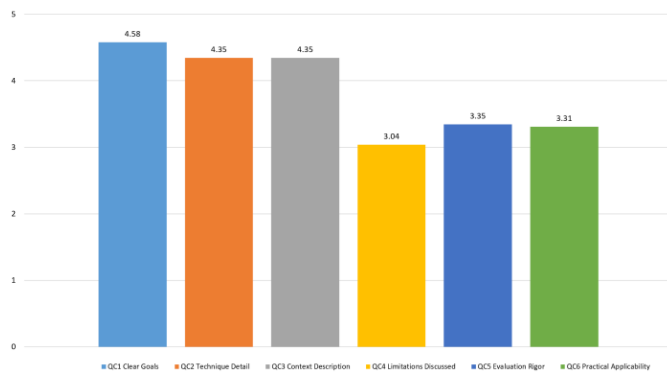

Fig. 9 Quality assessment of reviewed SLR publications.

## IV. RESEARCH DIRECTIONS AND FUTURE WORK

In this section the major challenges faced, future work and the research directions (RDs) for coverage-based V&V and safety assurance of SAVs are summarized based on the experience of conducting this SLR. They are as follows:

*RD1) Unification of terminologies*: Analysis of RQ2 showed us that approximately half of the papers did not explicitly mention the kind of coverage criteria they were using. This result revealed that many researchers around the world are pursuing research in this field but are using different terminologies for the same procedures. If the researchers in this field of V&V and safety assurance of SAVs can unite and agree on using the same set of terminologies for the same procedures, we believe the research in this domain would progress at a higher rate, also the dissemination and understanding of the research would go smoother.

*RD2) Explore new techniques for maximizing coverage*: RQ4 analysis shows us that mostly pseudorandom generation and SBST have been used to maximize coverage. This field is relatively new, and it is understandable, but we believe that numerous other techniques and their combinations e.g. different optimization techniques, ML, etc., can also be explored to maximize coverage for testing.

*RD3) Increase variety in test setup*: Fig. 8 depicting the results of RQ5 shows us that most of the research works (23 out of 26) rely on simulations, while only three publications involve hardware. Hardware should be involved frequently with the experimentations, as hardware can expose a lot of faults that a simulation cannot, as these simulations are generally low fidelity which was observed in the publications analysed in this SLR.

*RD4) Quality assessment observations*: From quality assessment of the reviewed SLR publications in Fig. 9, we can observe that the research works for coverage-based V&V and safety assurance of SAVs need to address the QCs that have relatively lower scores as highlighted in the previous section i.e., limitations (QC4), rigor in evaluation (QC5) and practical applicability (QC6) of their proposed methods.

## V. Conclusion

SAVs taking over the automobile market is looking more realistic day by day, the safety guarantee of SAVs is becoming a major concern. We have conducted a SLR in this domain of SAV safety assurance, on the topic of coverage-based testing for V&V and safety assurance of SAVs. The main objective of this SLR is to synthesize the existing body of knowledge about this topic and to identify research gaps and challenges for further research in this field.

This SLR draws out 26 articles in total related to the main topic of the SLR using a multi-stage process. The most pertinent findings of this SLR in terms of research gaps and challenges identification are as follows: (I) Non-standardization of terminologies; (II) More techniques to maximize coverage need to be explored; (III) Testing setup needs to get hardware involved more; (IV) Research works in this domain need to address limitations, rigor in evaluation and practical applicability of their proposed methods more in detail so that future works can improve upon those highlighted limitations. This SLR also classifies research in this domain on basis of their coverage criteria. Also, the list of all publications of this SLR mentioned with reference in Table IV as per technique used to maximize coverage, will help researchers in this domain to further develop and build upon these existing works to assure safety of SAVs in the future.

In future we plan to use the research works mentioned in this SLR and the presented research directions to investigate if we can develop even more effective coverage criteria than the existing ones, along with coverage maximizing techniques to assure safety and V&V of SAVs in an urban environment.

ACKNOWLEDGMENT

The research presented in this paper has been funded by European Union's EU Framework Programme for Research and Innovation Horizon 2020 under Grant Agreement No. 812.788.